\documentclass[conference]{IEEEtran}
\IEEEoverridecommandlockouts
\usepackage{cite}
\usepackage{amsmath,amssymb,amsfonts}
\usepackage{graphicx}
\usepackage{textcomp}
\usepackage{xcolor}
\usepackage{xspace}
\usepackage{xargs}% Use more than one optional parameter in a new commands
\usepackage{booktabs} % For formal tables
\usepackage{multirow}
\usepackage{colortbl}
\usepackage{comment}
\usepackage{bigints}
\usepackage{bbm} 
\usepackage{pifont}
\usepackage{algorithm}
\usepackage{algpseudocode}
\usepackage{multicol}
\usepackage{multirow} 
\usepackage{graphicx}
\usepackage{caption}
\usepackage{subcaption}
\usepackage{lipsum}
\usepackage{marginnote}

\newcommand{\eg}{e.g.\xspace}
\newcommand{\ie}{i.e.\xspace}

\def\BibTeX{{\rm B\kern-.05em{\sc i\kern-.025em b}\kern-.08em
    T\kern-.1667em\lower.7ex\hbox{E}\kern-.125emX}}

\begin{document}

\title{Shedding Light on VLN Robustness: \\ A Black-box Framework for  \\ Indoor Lighting-based Adversarial Attack
}

\author{\IEEEauthorblockN{Chenyang Li}
\IEEEauthorblockA{
% \textit{College of Computing and Data Science} \\
\textit{Nanyang Technological University}\\
Singapore, Singapore}
\and
\IEEEauthorblockN{Wenbing Tang}
\IEEEauthorblockA{
% \textit{College of Information Engineering} \\
\textit{Northwest A\&F University}\\
Yangling, Shaanxi, China}
\and
\IEEEauthorblockN{Yihao Huang}
\IEEEauthorblockA{
% \textit{School of Computing} \\
\textit{National University of Singapore}\\
Singapore, Singapore}
\and
\IEEEauthorblockN{Sinong Simon Zhan}
\IEEEauthorblockA{
% \textit{Department of Electrical and Computer Engineering } \\
\textit{Northwestern University}\\
Evanston, IL, USA}
\and
\IEEEauthorblockN{Ming Hu}
\IEEEauthorblockA{
% \textit{School of Computing and Information Systems} \\
\textit{Singapore Management University}\\
Singapore, Singapore}
\and
\IEEEauthorblockN{Xiaojun Jia}
\IEEEauthorblockA{
% \textit{College of Computing and Data Science} \\
\textit{Nanyang Technological University}\\
Singapore, Singapore}
\and
\IEEEauthorblockN{Yang Liu}
\IEEEauthorblockA{
% \textit{College of Computing and Data Science} \\
\textit{Nanyang Technological University}\\
Singapore, Singapore}
}

\maketitle

\begin{abstract}
Vision-and-Language Navigation (VLN) agents have made remarkable progress, but their robustness remains insufficiently studied. Existing adversarial evaluations often rely on perturbations that manifest as unusual textures rarely encountered in everyday indoor environments. Errors under such contrived conditions have limited practical relevance, as real-world agents are unlikely to encounter such artificial patterns. In this work, we focus on indoor lighting, an intrinsic yet largely overlooked scene attribute that strongly influences navigation. We propose Indoor Lighting-based Adversarial Attack (ILA), a black-box framework that manipulates global illumination to disrupt VLN agents. Motivated by typical household lighting usage, we design two attack modes: Static Indoor Lighting-based Attack (SILA), where the lighting intensity remains constant throughout an episode, and Dynamic Indoor Lighting-based Attack (DILA), where lights are switched on or off at critical moments to induce abrupt illumination changes. We evaluate ILA on two state-of-the-art VLN models across three navigation tasks. Results show that ILA significantly increases failure rates while reducing trajectory efficiency, revealing previously unrecognized vulnerabilities of VLN agents to realistic indoor lighting variations.
\end{abstract}

% \begin{IEEEkeywords}
% component, formatting, style, styling, insert
% \end{IEEEkeywords}

% %%%%%%%%%%%%%%%%%%%%%%%%%%%%%%%%%%%%%%%%%%%%%%%%%%%%%%%%%%%%%%%%%%%%%%%%%%%%%%%%
% \begin{figure}[tb]
%     \centering
%     \includegraphics[width=1.0\linewidth]{figures/concept-5.pdf}
%     \caption{We propose Indoor Lighting-based adversarial Attack to evaluate the robustness of VLN models by a common approach.
%     }
%     \label{fig:concept}
% \end{figure}

% %%%%%%%%%%%%%%%%%%%%%%%%%%%%%%%%%%%%%%%%%%%%%%%%%%%%%%%%%%%%%%%%%%%%%%%%%%%%%%%%

\section{Introduction}
\label{sec:intro}

Embodied AI aims to create agents that complete complex tasks by interacting with the physical environment~\cite{liu2025embodiedai}. 
As a fundamental capability for such agents, Vision-and-Language Navigation (VLN) requires an agent to navigate through dynamic physical environments by following natural language instructions~\cite{anderson2018vlnr2r, thomason2020vlndialog, chen2019vlntouchdown}. 
Given that the majority of human-robot interactions occur in indoor scenarios, indoor VLN has become the primary research focus in the field~\cite{anderson2018vlnr2r, krantz2020vlnce, shridhar2020vlnalfred}. Although recent advances in models and training techniques have steadily improved VLN performance~\cite{zhang2024vlnsurveymodel, wu2024embodiednavigationsurveymodel, gu2022vlnsurvey}, the robustness of VLN models remains insufficiently explored. This gap is critical, as navigation failures could lead to collisions, injuries, or other hazardous consequences~\cite{denning2009householdsafety, mitka2012householdsafety, tadele2014householdsafety}.

Recent studies have examined the robustness of VLN models through adversarial attacks~\cite{goodfellow2014whitefgsm,madry2017whitepgd}, which manipulate visual perception to mislead navigation. These works~\cite{yang2025hijacking,chen2024physicalizable} embed adversarial patterns into environments by altering objects' textures or adding patches on objects' surfaces, to achieve strong attack performance.
However, the perturbations produced by these methods often manifest as highly unusual textures that are rarely encountered in everyday indoor settings. Errors under such contrived conditions have limited practical relevance, since the likelihood of encountering such artificial patterns in real-world scenarios is extremely low. For agents, however, it is essential to examine vulnerabilities arising from natural and unavoidable variations in the environment. To this end, we focus on intrinsic scene attributes that pervade daily indoor life. In particular, we highlight indoor lighting as an essential and variable element of indoor environments that profoundly shapes the agent’s perception and plays a critical role in evaluating the robustness of VLN models. 

In this paper, we propose a black-box \textbf{I}ndoor \textbf{L}ighting-based adversarial \textbf{A}ttack, termed \textbf{ILA}. Motivated by common usage patterns of household lighting, where lights are often kept at a steady level but may also be abruptly switched on or off, we design two attack modes: a \textbf{Static} indoor lighting-based attack (\textbf{SILA}), which searches for globally disruptive lighting configurations, and a \textbf{Dynamic} indoor lighting-based attack (\textbf{DILA}), which triggers sudden illumination changes at critical decision points. 
For the sake of convenience, we assume control of a single global light and that each black-box query returns the agent’s position and movement heading.
SILA searches candidate global intensities, compares the resulting trajectories, and picks the intensity that maximizes trajectory deviation. 
Based on the intensity achieved by SILA, DILA queries the agent at each step to estimate how switching the light on or off would alter the agent's heading, and triggers a light switch when doing so increases the deviation from the navigation goal.

In summary, our contributions are as follows:
\begin{itemize}
    \item We introduce a black-box adversarial paradigm that perturbs intrinsic environmental attributes, specifically indoor lighting intensity, as a practical method for evaluating the robustness and reliability of indoor VLN models.
    \item We propose two attack modes inspired by household lighting. SILA searches global intensities to expose trajectory-level sensitivity to steady illumination changes. DILA triggers abrupt on/off switches based on the agent’s state to provoke unsuitable action selection.
    \item We conduct experiments on two SOTA VLN models across three tasks with a total of 576 episodes, demonstrating that our method substantially reduces the VLN performance of the agent.
\end{itemize}

%%%%%%%%%%%%%%%%%%%%%%%%%%%%%%%%%%%%%%%%%%%%%%%%%%%%%%%%%%%%%%%%%%%%%%%%%%%%%%%%%%%%%%%%%%%%%%%%%%%%%%%%
\section{Related Work}
\label{sec:related_work}

\textbf{Vision-and-Language Navigation (VLN)}. 
VLN~\cite{anderson2018vlnr2r} tasks require agents to follow natural language instructions and perceive their surroundings to navigate unseen environments. Successful navigation depends on integrating visual perception, language understanding, and action planning. VLN tasks fall into outdoor~\cite{chen2019vlntouchdown, sriram2019vlnlcsd, liu2023vlnaerial} and indoor~\cite{anderson2018vlnr2r, krantz2020vlnce, shridhar2020vlnalfred} categories. This work focuses on indoor VLN, which involves embodied navigation in structured 3D environments~\cite{sulaiman2020matterport, savva2019habitat, kolve2017ai2thor},
where agents use egocentric cameras and basic locomotion.
Models for indoor VLN have evolved from sequence-to-sequence architectures~\cite{anderson2018vlnr2r} with CNN-LSTM encoders and attention, to Transformer-based methods~\cite{hao2020prevalent, chen2021hamt, ehsani2024spoc, hu2025flare} that better capture complex language and spatial relations.
Combined with RL, IL, and pretraining, these models achieve strong navigation performance. 
For instance, SPOC~\cite{ehsani2024spoc} employs behavior cloning on shortest-path expert trajectories, while FLaRe~\cite{hu2025flare} extends SPOC with RL fine-tuning.

\textbf{Adversarial Attacks}.
Adversarial attacks add human-imperceptible perturbations to inputs to mislead models~\cite{goodfellow2014whitefgsm}, causing them to fail despite performing well on normal samples, thereby exposing robustness deficiencies.
These attacks have been widely studied across various domains, including vision~\cite{goodfellow2014whitefgsm}, speech~\cite{carlini2018audio}, language~\cite{gao2018text}, and autonomous driving~\cite{cao2019av}, serving as a critical tool for robustness evaluation.
Adversarial attacks can be classified into white-box and black-box attacks depending on the level of access to the target model.
White-box attacks assume full access to model architecture, weights, and gradients, and craft perturbations by computing gradients with respect to the input and iteratively optimizing adversarial objectives.
Representative methods include Fast Gradient Sign Method~\cite{goodfellow2014whitefgsm}, Projected Gradient Descent~\cite{madry2017whitepgd}, and the Carlini \& Wagner~\cite{carlini2017whitecw} attack.
In contrast, black-box attacks operate without access to model internals, employing gradient-free search strategies that iteratively refine candidate perturbations based on queried outputs. 
Representative methods include Natural Evolution Strategies~\cite{ilyas2018blacknes}, Bandits-TD~\cite{ilyas2018blackbandits}, and SimBA~\cite{guo2019blacksimba}.

\textbf{VLN Robustness Evaluation}.
Assessing how VLN models respond to perceptual variations is critical for ensuring reliable deployment. 
Recent work has investigated adversarial attacks on embodied navigation agents by perturbing visual inputs or altering environmental attributes. Consistent Attack~\cite{ying2023uap} applies universal adversarial perturbations to continuously influence agent perception. Yang et al.\cite{yang2025hijacking} optimize object appearances via differentiable rendering to manipulate agent behavior. Chen et al.\cite{chen2024physicalizable} apply adversarial patches to objects, optimizing textures across multiple viewpoints to alter agent decisions.
However, these approaches share common limitations: First, the perturbations often produce highly unusual textures rarely encountered in everyday indoor settings, reducing their practical relevance. Second, they rely on white-box access to model internals for optimization, limiting their applicability in realistic threat scenarios where such access is unavailable.

%%%%%%%%%%%%%%%%%%%%%%%%%%%%%%%%%%%%%%%%%%%%%%%%%%%%%%%%%%%%%%%%%%%%%%%%%%%%%%%%%%%%%%%%%%%%%%
\begin{figure}[tb]
    \centering
    \includegraphics[width=1.0\linewidth]{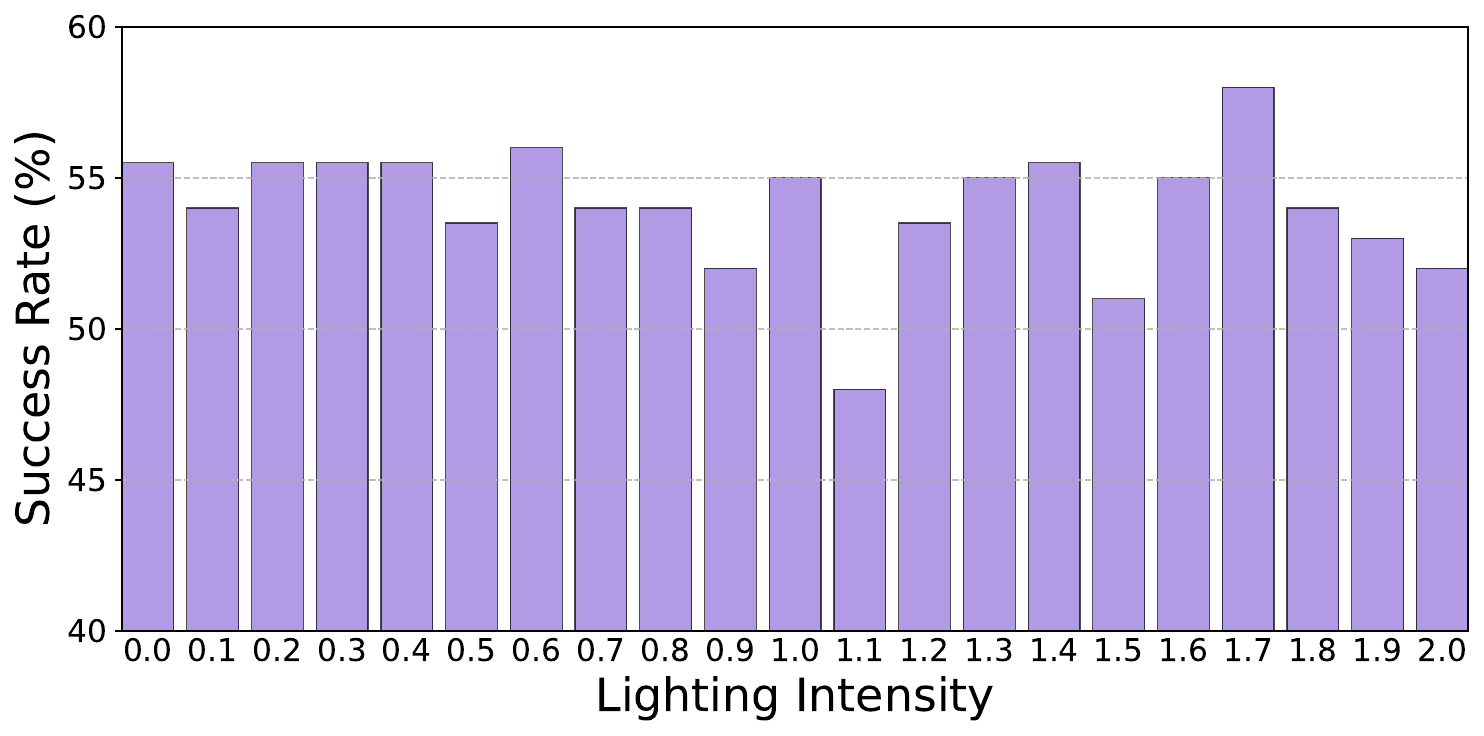}
    \caption{
    Impact of lighting intensity on navigation success rate.
    }
    \label{fig:motivation}
\end{figure}
%%%%%%%%%%%%%%%%%%%%%%%%%%%%%%%%%%%%%%%%%%%%%%%%%%%%%%%%%%%%%%%%%%%%%%%%%%%%%%%%%%%%%%%%%%%%%%

%%%%%%%%%%%%%%%%%%%%%%%%%%%%%%%%%%%%%%%%%%%%%%%%%%%%%%%%%%%%%%%%%%%%%%%%%%%%%%%%%%%%%%%%%%%%%%
\section{Preliminary}
\label{sec:motivation}
\subsection{Motivation}

Motivated by the need for more realistic robustness evaluation, we conduct an empirical study to investigate whether indoor lighting, a naturally occurring and easily controlled environmental factor, can affect VLN agent performance. 
Specifically, we conduct experiments using a VLN model from SPOC~\cite{ehsani2024spoc} on the Object Navigation task of the CHORES dataset~\cite{ehsani2024spoc}. In this task, the agent is required to navigate to a specified object category in an unseen environment. The evaluation included 200 episodes, during which we systematically varied the lighting intensity within the range [0,2], with increments of 0.1.
The results are shown in Figure~\ref{fig:motivation}, where the x-axis denotes lighting intensity and the y-axis represents task success rate. 
From the figure, we observe two key patterns: 
\ding{182} Distinct and irregular performance fluctuations are observed across lighting intensities within the normal range. The success rate shows no monotonic trend, indicating that illumination changes impact the model’s behavior in a nonlinear manner. \ding{183} The performance gap between the highest and lowest success rates approaches ten percentage points. These observations suggest that even moderate variations in lighting can affect the navigation behavior of VLN agents, raising an important question: 
\textit{Can intentionally designed lighting modulation patterns systematically expose the potential vulnerabilities of VLN models to illumination changes?}

\subsection{Problem Definition}
\label{ssec: problem_definition}
Let $E$ denote an indoor environment and $s_t$ the agent's state (pose) at timestep $t=1,\dots,T$, where $T$ is the maximum number of timesteps.
An episode represents a single navigation case, starting from the initial state $s_1$ with instruction $I$ that specifies the target destination $G$, and terminating when the agent issues STOP (indicating it believes the goal is reached) or when $T$ is reached.
The indoor lighting is represented by a time-varying global intensity sequence $\mathcal{L}=\{l_1,\dots,l_T\}$, where $l_t$ is the intensity at step $t$. 

A rendering function $r$ maps the environment $E$, the agent state $s_t$, and the current lighting $l_t$ to the agent’s current visual observation:
\begin{equation}
o_t = r(E, s_t, l_t).
\end{equation}
With the natural-language instruction $I$, the agent follows a policy $\pi$ and selects actions
\begin{equation}
a_t \sim \pi(a_t \mid o_{1:t}, I),
\end{equation}
which transitions the agent to a new state $s_{t+1}$ with a state-transition function $\Phi$:
\begin{equation}
s_{t+1} = \Phi(s_t, a_t).
\end{equation}
The agent thus produces a trajectory $\tau = \{s_1, s_2, \dots, s_{\hat{T}}\}$, where $\hat{T}$ denotes the actual timesteps of an episode.

Thus, we define the VLN model $V(\cdot)$ of the agent as
\begin{equation}
\tau = V(E, \mathcal{L}, I, s_1).
\end{equation}

\noindent \textbf{Objective.} 
The attacker's objective is to find a lighting condition that causes the agent's navigation episode to fail. Thus, we define the evaluation result
\begin{equation}
    \mathcal{R} = \Theta(\tau, I),
\end{equation}
where $\Theta(\cdot)$ is the evaluation function, and an episode is considered \emph{successful} if $\mathcal{R}$ is true and \emph{failure} otherwise. 

To express this objective more concretely, we further define the success and failure conditions at the action level.
Let $G$ denote the goal region (\ie, destination) and $\mathcal{L}$ the controllable lighting sequence. 
An episode is considered \emph{successful} if there exists a timestep $t\le T$ such that $\operatorname{pos}(s_t)\in G$ and $a_t=\text{STOP}$. 
The attacker aims to induce \emph{failure} episodes, \ie, episodes in which the agent never satisfies this success condition. 
Formally, the objective is
\begin{equation}
\small
\max_{\mathcal{L}}\; \mathbf{1}\!\left(\forall\, t\le T,\;\neg\big(\operatorname{pos}(s_t)\in G \land a_t=\text{STOP}\big)\right),
\end{equation}
where $\mathbf{1}(\cdot)$ is the indicator function, $\operatorname{pos}(\cdot)$ extracts the position component of state $s_t$, and $a_t$ is the action taken at step $t$ in the trajectory $\tau$ induced by $\mathcal{L}$.
%%%%%%%%%%%%%%%%%%%%%%%%%%%%%%%%%%%%%%%%%%%%%%%%%%%%%%%%%%%%%%%%%%%%%%

%%%%%%%%%%%%%%%%%%%%%%%%%%%%%%%%%%%%%%%%%%%%%%%%%%%%%%%%%%%%%%%%%%%%%%
\begin{figure*}[t]
    \centering
    \includegraphics[width=\linewidth]{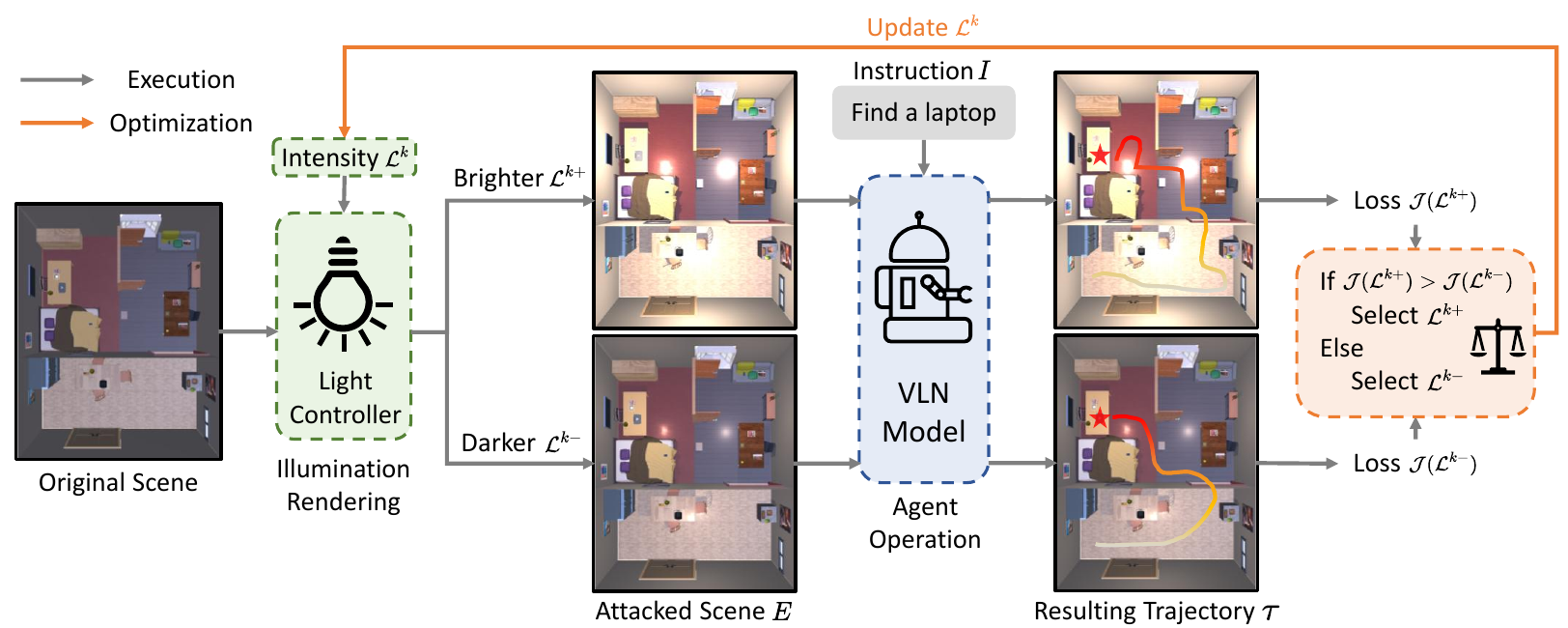}
    \caption{
        Pipeline of Static Indoor Lighting-based Attack.
        The light controller renders two perturbed illumination conditions, and then the agent performs navigation under each condition, producing corresponding trajectories and losses. The loss comparison determines which illumination yields stronger disruption, and the selected intensity is used to update the lighting configuration for the next iteration.
    }
    \label{fig:static}
\end{figure*}

%%%%%%%%%%%%%%%%%%%%%%%%%%%%%%%%%%%%%%%%%%%%%%%%%%%%%%%%%%%%%%%

%%%%%%%%%%%%%%%%%%%%%%%%%%%%%%%%%%%%%%%%%%%%%%%%%%%%%%%%%%%%%%%%%%%%%%%%%%%%%%%%%%%
\section{Method}
\label{sec:method}
\subsection{Overview}
\label{ssec:overview}
% From the problem formulation above, 
Based on the above problem definition, the attacker’s only controllable variable is the lighting sequence. 
We design our method around two everyday indoor lighting scenarios: a static scenario, where lighting intensity remains constant throughout the episode, and a dynamic scenario, where lights are switched on or off at particular moments, producing abrupt illumination changes.
Accordingly, we introduce two attack modes: a static indoor lighting-based attack (SILA), which searches for a unified globally disruptive lighting configuration, and a dynamic indoor lighting-based attack (DILA), which induces sudden illumination changes at critical decision points to destabilize the agent. These two modes help reveal VLN agents' vulnerability under typical indoor lighting conditions. 
We detail SILA and DILA in Sections~\ref{ssec:static} and~\ref{ssec:dynamic}, respectively.

%%%%%%%%%%%%%%%%%%%%%%%%%%%%%%%%%%%%%%%%%%%%%%%%%%%%%%%%%%%%%%%%%%%%%%%%%%%%%%%%%%%

\subsection{Static Indoor Lighting-based Attack}
\label{ssec:static}
We use a black-box adversarial attack to search for the adversarial lighting intensity. Before introducing the loss function and optimization procedure, we first outline the overall attack pipeline, as illustrated in Figure~\ref{fig:static}. Each iteration follows a closed loop: the lighting controller selects a candidate global intensity, the environment is rendered accordingly, and the agent executes navigation. The resulting trajectory is evaluated by a timestep-weighted loss, which provides feedback to guide the next update. This loop forms the foundation of our adversarial optimization.

The lighting is fixed for the entire episode. 
Concretely, the lighting sequence is restricted to be
\begin{equation}
\mathcal{L}_{\text{static}}=\{l^\star,\dots,l^\star\}, \qquad l^\star\in[l_{\min},l_{\max}]. 
\end{equation}
This represents a constant illumination setting (\eg, always bright) intended to globally degrade perception.

\subsubsection{Timestep-weighted Loss Design}
Designing an attack loss for navigation tasks is challenging. Unlike classification tasks that output a single decision, navigation unfolds as a sequential process, producing an entire motion trajectory. Thus, defining loss only by whether the agent finally reaches the goal discards rich trajectory information. 
To address this, we propose to design the loss over the whole trajectory. Moreover, early deviations can often be corrected, while late-stage errors are usually decisive and irreversible. 
Therefore, treating all timesteps equally misrepresents their true importance. To capture this asymmetry, we design the loss with timestep-increasing weights, emphasizing errors closer to the end of the episode.

Given $\mathcal{L}_{\text{static}}$ as the static lighting sequence, the trajectory $\tau_{\text{static}} = V(E, \mathcal{L}_{\text{static}}, I, s_1)$. $\hat{T} = |\tau|$ denotes the actual execution timesteps of an episode. 
We set increasing weights
\begin{equation}
w_i = \frac{t}{\hat{T}}, \qquad t=1,\dots,\hat{T}.
\end{equation}

For the trajectory $\tau_{\text{static}}=\{s_1,\dots,s_{\hat{T}}\}$, the agent position is $p_t=\operatorname{pos}(s_t)$. We define a per-step deviation $d_t = \|p_t - G\|_2$ as the Euclidean distance between the agent’s position $p_t$ and $G$. The timestep-weighted loss for the navigation task is
\begin{align}
\mathcal{J}_{\text{static}}= \sum_{t=1}^{\hat{T}} w_t\, d_t = \sum_{t=1}^{\hat{T}} \frac{t}{\hat{T}} \|\operatorname{pos}(s_t) - G\|_2.
\end{align}

\subsubsection{Optimization of Lighting Intensity} 
We treat the search for adversarial lighting intensity as a black-box optimization problem, where the task loss provides feedback for adjusting the lighting. At each iteration, we test two nearby lighting settings, one slightly brighter and one slightly darker than the current intensity, compare their task losses, and then choose the setting that produces the larger loss, meaning the one that more strongly disrupts the agent’s navigation trajectory.

Let $l^0$ denote the initial lighting intensity, $\Delta l$ the accumulated intensity offset (initialized as $\Delta l=0$), and $\alpha>0$ the perturbation step size of lighting intensity. At iteration $k\in\{1,\dots,K\}$ (where $K$ is the total number of attack iterations), the current intensity is $l^k = l^0 + \Delta l$. 
We evaluate two illumination candidates:
\begin{equation}
l^{k+} = \operatorname{clip}(l^k+\alpha), \qquad
l^{k-} = \operatorname{clip}(l^k-\alpha),
\end{equation}
where $\operatorname{clip}(\cdot)$ projects values into the admissible interval $[l_{\min},\,l_{\max}]$. 
These two conditions define two lighting sequences $\mathcal{L}^{k+} \!=\! \{l^{k+}, \dots, l^{k+}\}$ and $\mathcal{L}^{k-} \!=\! \{l^{k-}, \dots, l^{k-}\}$.
We run the VLN model under each sequence to obtain trajectories, 
from which we compute the corresponding losses 
$\mathcal{J}(\mathcal{L}^{k+})$ and $\mathcal{J}(\mathcal{L}^{k-})$,  respectively. 
Define
\begin{equation}
\xi^k = \operatorname{sign}(\mathcal{J}(\mathcal{L}^{k+}) - \mathcal{J}(\mathcal{L}^{k-})) \in \{+1,-1\},
\end{equation}
so that $\xi^k=+1$ indicates $\mathcal{L}^{k+}$ is worse (larger loss) and $\xi^k=-1$ indicates $\mathcal{L}^{k-}$ is worse.

To avoid local stagnation, we adopt an $\varepsilon$-greedy strategy that occasionally reverses the update direction. 
Sampling $u\sim\mathrm{Uniform}(0,1)$ with exploration rate $\varepsilon\in[0,1]$, we set
\begin{equation}
b^k = 
\begin{cases}
+1, & u \ge \varepsilon, \\[2pt]
-1, & u < \varepsilon,
\end{cases}
\end{equation}
where $b^k=+1$ follows the worse direction and $b^k=-1$ flips once for exploration. 
The update then becomes
\begin{equation}
\Delta l \leftarrow \operatorname{clip}\!\bigl(\Delta l+\alpha\cdot b^k \cdot \xi^k \bigr),
\end{equation}
\begin{equation}
l^\star = l^0 + \Delta l.
\end{equation}

Through two-sided comparison and $\varepsilon$-greedy exploration, the search is directed toward lighting intensities that maximize loss, with termination triggered by task failure or reaching the iteration limit.
The complete SILA algorithm is presented in Algorithm~\ref{alg:static_attack}.

%%%%%%%%%%%%%%%%%%%%%%%%%%%%%%%%%%%%%%%%%%%%%%%%%%%%%%%%%%%%%%%%%%%%%%%%%%%
\begin{algorithm}[tbp]
\caption{Static Indoor Lighting-based Attack.}
\label{alg:static_attack}
\textbf{Input:} Initial light intensity $l^0$, accumulated intensity offset $\Delta l$, per-iteration step size $\alpha$, number of iterations $K$, environment $E$, instruction $I$, initial state $s_1$, lighting sequence $\mathcal{L} = \{l^\star, \dots, l^\star\}$\\
\textbf{Output:} Adversarial lighting intensity $l^\star$

\begin{algorithmic}[1]
\State $l^\star \gets l^0$,  $\Delta l \gets 0$ 
\State $\tau^0 \gets V(E, \mathcal{L}^{0}, I, s_1)$
\If{$ \neg \big(\Theta(\tau^0, I)\big)$}
    \State \Return $l^\star \gets l^0$
\EndIf
\For{$k = 1$ to $K$}
    \State $l^k \gets l^0+\Delta l$
    \State $l^{k+} \!\gets \operatorname{clip}(l^k+\alpha)$, $l^{k-} \!\gets \operatorname{clip}(l^k-\alpha)$
    \State $\tau^{k+} \!\gets V(E, \mathcal{L}^{k+}\!, I, s_1), \tau^{k-} \!\gets V(E, \mathcal{L}^{k-}\!, I, s_1)$
    \If{$\neg \big(\Theta(\tau^{k+}, I)\big)$}
        \State return $l^\star \gets l^{k+}$
    \EndIf
    \If{$\neg \big(\Theta(\tau^{k-}, I)\big)$}
        \State return $l^\star \gets l^{k-}$
    \EndIf
    \State $\mathcal{J}(\mathcal{L}^{k+}) \gets \{\tau^{k+}, I\}, \mathcal{J}(\mathcal{L}^{k-}) \gets \{\tau^{k-}, I\}$
    \State $\xi^k \gets \operatorname{sign}(\mathcal{J}(\mathcal{L}^{k+})\!-\!\mathcal{J}(\mathcal{L}^{k-}))$
    \State Sample $b^k$
    \State $\Delta l \gets \operatorname{clip}\bigl(\Delta l+\alpha \cdot b^k \cdot \xi^k\bigr)$
    \State $l^\star \gets l^0 + \Delta l$
\EndFor
\end{algorithmic}
\end{algorithm}
%%%%%%%%%%%%%%%%%%%%%%%%%%%%%%%%%%%%%%%%%%%%%%%%%%%%%%%%%%

%%%%%%%%%%%%%%%%%%%%%%%%%%%%%%%%%%%%%%%%%%%%%%%%%%%%%%%%%%%%%%%
\begin{figure}[t]
    \centering
    \begin{subfigure}[t]{\linewidth}
        \centering
        \includegraphics[width=\linewidth]{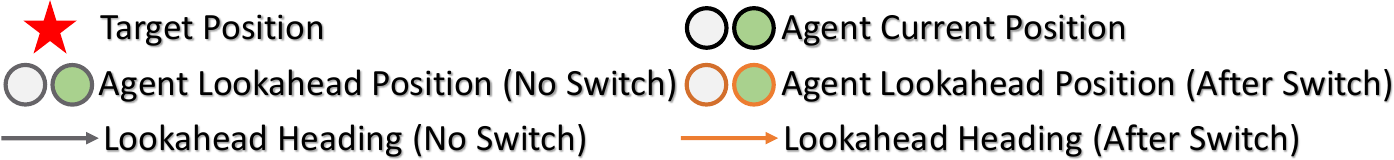}
    \end{subfigure}
    
    \begin{subfigure}{0.49\linewidth}
        \centering
        \includegraphics[width=\linewidth]{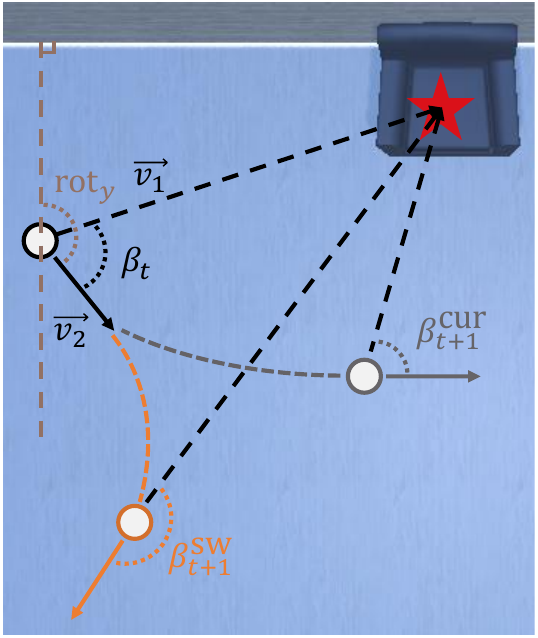}
        \caption{Deviation Increase}
        \label{fig:dynamic_increase}
    \end{subfigure}
    \begin{subfigure}{0.49\linewidth}
        \centering
        \includegraphics[width=\linewidth]{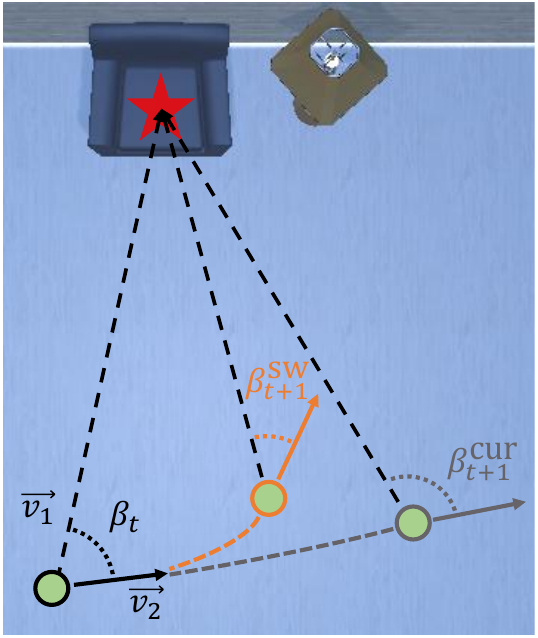}
        \caption{Deviation Decrease}
        \label{fig:dynamic_decrease}
    \end{subfigure}
    \caption{
        Illustration of Dynamic Indoor Lighting-based Attack. 
        The switch is triggered when the deviation of the angle between the agent's toward-target direction and its heading increases (left), and remains unchanged when the deviation decreases (right).
    }
    \label{fig:dynamic}
\end{figure}
%%%%%%%%%%%%%%%%%%%%%%%%%%%%%%%%%%%%%%%%%%%%%%%%%

\subsection{Dynamic Indoor Lighting-based Attack}
\label{ssec:dynamic}

In the dynamic attack mode, the only difference from the static setting is the lighting sequence. Here we restrict lighting to an on/off pattern with a fixed on-level $l^\star\in[l_{\min},l_{\max}]$ and off-level $0$. Let $i_t\in\{0,1\}$ be the on/off indicator at timestep $t$. Then the per-step intensity is
\begin{equation}
l_t = i_t \cdot l^\star,\qquad t=1,\dots,\hat{T}.
\end{equation}
Note that we set the light intensity to be $l^\star$ obtained from the static mode, using it as the default ``on'' brightness in the dynamic setting. This ensures that the dynamic attack inherits the strongest static degradation effect while further exploiting temporal switching to disrupt navigation.

Therefore, in the dynamic attack, our goal is to select suitable timesteps for switching the light on or off. This selection faces two main challenges. \ding{182} Waiting for the full trajectory to evaluate every candidate switch is slow and resource-intensive. \ding{183} When multiple switches occur in one episode, it becomes difficult to attribute which particular switch(s) caused the failure.
To address these issues, we design a lightweight one-step lookahead surrogate that estimates the immediate impact of a single switch and uses it to decide whether to switch, thereby avoiding costly full-rollout evaluations while still capturing action signals for effective switching. Specifically, as shown in Figure~\ref{fig:dynamic}, let $p_t=(\mathrm{pos}_{x},\mathrm{pos}_{z})$ denote the agent's position at timestep $t$ (where $y$ denotes the vertical axis), $G=(\mathrm{tar}_{x},\mathrm{tar}_{z})$ the target location, and  $\mathrm{rot}_{y}$ the heading angle. 
We define the toward-target vector $\vec v_1$ and the facing vector $\vec v_2$ as
\begin{equation}
\vec v_1=(\mathrm{tar}_{x}-\mathrm{pos}_{x},\ \mathrm{tar}_{z}-\mathrm{pos}_{z}),
\end{equation}
\begin{equation}
\vec v_2=(\sin(\mathrm{rot}_{y}),\ \cos(\mathrm{rot}_{y})),
\end{equation}
and compute their relative angle
\begin{equation}
\beta_t=\arccos\!\left(\frac{\vec v_1\cdot\vec v_2}{\|\vec v_1\|_2\,\|\vec v_2\|_2}\right).
\end{equation}

Based on $\beta_t$, we evaluate whether a lighting switch would increase deviation.
Specifically, we perform a one-step lookahead under both the current lighting $l_t$ and the switched lighting $\tilde l_t$, 
feed the resulting observations into the VLN model to obtain actions $a_t^{\mathrm{cur}}$ and $a_t^{\mathrm{sw}}$, 
and simulate the next states $s_{t+1}^\text{cur}$ and $s_{t+1}^\text{sw}$. Then, we compute $\beta_{t+1}^{\mathrm{cur}}$ and $\beta_{t+1}^{\mathrm{sw}}$ based on these two states, and
a switch is triggered if
\begin{equation}
\beta_{t+1}^{\mathrm{sw}} - \beta_{t+1}^{\mathrm{cur}} > 0,
\end{equation}
% \ie, when switching increases the deviation from the target. 
\ie, when switching amplifies the deviation between the agent’s heading and the toward-target direction to steer the agent further away from the goal gradually, as shown in Figure~\ref{fig:dynamic_increase}. Otherwise, the current lighting on/off is kept if the deviation decreases, as shown in Figure~\ref{fig:dynamic_decrease}.

%%%%%%%%%%%%%%%%%%%%%%%%%%%%%%%%%%%%%%%%%%%%%%%%%%%%%%%%%%%%%%%%%%%%%%%%%%%

\begin{table*}[t]
\centering
\caption{Attack performance on SPOC and FLaRe over ObjectNav, Fetch, and RoomVisit tasks. 
}
\label{tab:main_result}
\resizebox{\linewidth}{!}{
\begin{tabular}{|l|cc|cc|cc|cc|cc|cc|}
\hline
\textbf{Model} 
& \multicolumn{6}{c|}{\textbf{SPOC}} & \multicolumn{6}{c|}{\textbf{FLaRe}} \\
\cline{2-7} \cline{8-13}
\textbf{Task} & \multicolumn{2}{c|}{\textit{ObjectNav}} & \multicolumn{2}{c|}{\textit{Fetch}} & \multicolumn{2}{c|}{\textit{RoomVisit}} 
& \multicolumn{2}{c|}{\textit{ObjectNav}} & \multicolumn{2}{c|}{\textit{Fetch}} & \multicolumn{2}{c|}{\textit{RoomVisit}}  \\
\cline{2-3} \cline{4-5} \cline{6-7}
\cline{8-9} \cline{10-11} \cline{12-13}
\textbf{Metric} & ASR$\uparrow$ & EL$\uparrow$ & ASR$\uparrow$ & EL$\uparrow$ & ASR$\uparrow$ & EL$\uparrow$ 
& ASR$\uparrow$ & EL$\uparrow$ & ASR$\uparrow$ & EL$\uparrow$ & ASR$\uparrow$ & EL$\uparrow$  \\
\hline
No Attack    & 0.00 & 115.35 & 0.00 & 344.66 & 0.00 & 304.07  & 0.00 & 104.07 & 0.00 & 236.79 & 0.00 & 196.43 \\
Random Intensity      & 23.23 & 129.12 & 75.00 & 371.63 & 23.75 & 309.49  & 12.20 & 121.65 & 17.27 & 260.38 & 15.08 & 186.70 \\ 
Texture-GA & 54.90 & 137.04 & 87.50 & 386.72 & 50.62 & 327.44   & 37.57 & 149.62 & 53.98 & 311.46 & 45.53 & 192.95\\
\hline
Ours(only SILA)     & 60.38 & 131.07 & \textbf{100.00} & 367.97 & 52.44 & 345.50  & 47.27 & 153.86 & 68.81 & 327.21 & 57.48 & 181.16 \\
Ours(SILA+DILA)     & \textbf{96.23} & 234.27 & \textbf{100.00} & 367.97 & \textbf{70.73} & 369.99  & \textbf{52.73} & 185.84 & \textbf{93.58} & 409.84 & \textbf{74.80} & 187.89 \\
\hline
\end{tabular}
}
\end{table*}
%%%%%%%%%%%%%%%%%%%%%%%%%%%%%%%%%%%%%%%%%%%%%%%%%%%%%%%%%%%%%%%%%

\section{Experiments}
\label{sec:experiments}

\subsection{Experiment Setup}
\label{ssec:exp_set}

\textbf{Models, tasks \& environment.} We evaluate our method on two SOTA VLN models, SPOC~\cite{ehsani2024spoc} and FLaRe~\cite{hu2025flare}. We adopt the official pretrained checkpoints without any fine-tuning and follow the standard inference protocols. We use the CHORES~\cite{ehsani2024spoc} benchmark, which includes three indoor navigation-related tasks: ObjectNav, Fetch, and RoomVisit. We report results on their official test splits with 200, 200, and 172 episodes, respectively. In ObjectNav, the agent needs to locate and navigate to a specified object category. Fetch extends this by requiring the agent to pick up the object after reaching it. RoomVisit requires the agent to visit all distinct rooms in the scene. 
All scenes are rendered in the AI2-THOR~\cite{kolve2017ai2thor} simulator.

\noindent\textbf{Baselines.}  To evaluate our method, we compare it against two representative baselines. \textit{Random Light Intensity} evaluates the effect of random lighting variations on VLN models. For each episode, an intensity is uniformly sampled from the valid range and remains constant throughout the episode. 
\textit{Texture-GA} implements a black-box variant of texture-based attacks~\cite{chen2024physicalizable} that perturbs wall textures. For each episode, wall textures remain fixed during one execution and are optimized iteratively using a genetic algorithm, while all other scene components remain unchanged.

\noindent\textbf{Evaluation metrics.} 
We evaluate using two metrics: Attack Success Rate (ASR) and Episode Length (EL). ASR measures the fraction of episodes that succeed in the clean environment but fail under attack, directly reflecting attack effectiveness. EL measures the average number of timesteps taken by the agent during an episode, with longer episodes indicating less efficient navigation.
Formally, let $\mathcal{E}_{\text{task}}$ denote the set of episodes for a given task. For each episode $e\in\mathcal{E}_{\text{task}}$, let $\eta_{\text{clean}}(e),\; \eta_{\text{attack}}(e)\in\{1,0\}$ indicate success or failure for clean and attacked conditions, respectively. We define ASR over the subset of episodes that succeed in the clean setting:
\begin{equation}
\mathcal{E}_{\text{task}}^{1} =\{\,e\in\mathcal{E}_{\text{task}}\mid \eta_{\text{clean}}(e)=1\,\},
\end{equation}
\begin{equation}
\mathrm{ASR}=\frac{1}{|\mathcal{E}_{\text{task}}^{1}|}\sum_{e\in \mathcal{E}_{\text{task}}^{1}}\mathbbm{1}\!\big[\eta_{\text{attack}}(e)=0\big].
\end{equation}
For EL, let $|\tau(e)|$ denote the number of steps taken in episode $e$. We compute EL as the average episode length:
\begin{equation}
\mathrm{EL}=\frac{1}{|\mathcal{E}_{\text{task}}|}\sum_{e\in\mathcal{E}_{\text{task}}} |\tau(e)|.
\end{equation}
%%%%%%%%%%%%%%%%%%%%%%%%%%%%%%%%%%%%%%%%%%%%%%%%%%%%%%%%%%%%%%%%%

%%%%%%%%%%%%%%%%%%%%%%%%%%%%%%%%%%%%%%%%%%%%%%%%%%%%%%%%%%%%%%%%%%%%%%
\begin{figure*}[t]
    \begin{subfigure}{0.05\linewidth}
        \centering
        \includegraphics[width=\linewidth]{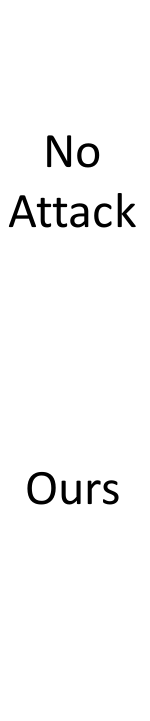}
    \end{subfigure}
    \begin{subfigure}{0.311\linewidth}
        \centering
        \includegraphics[width=\linewidth]{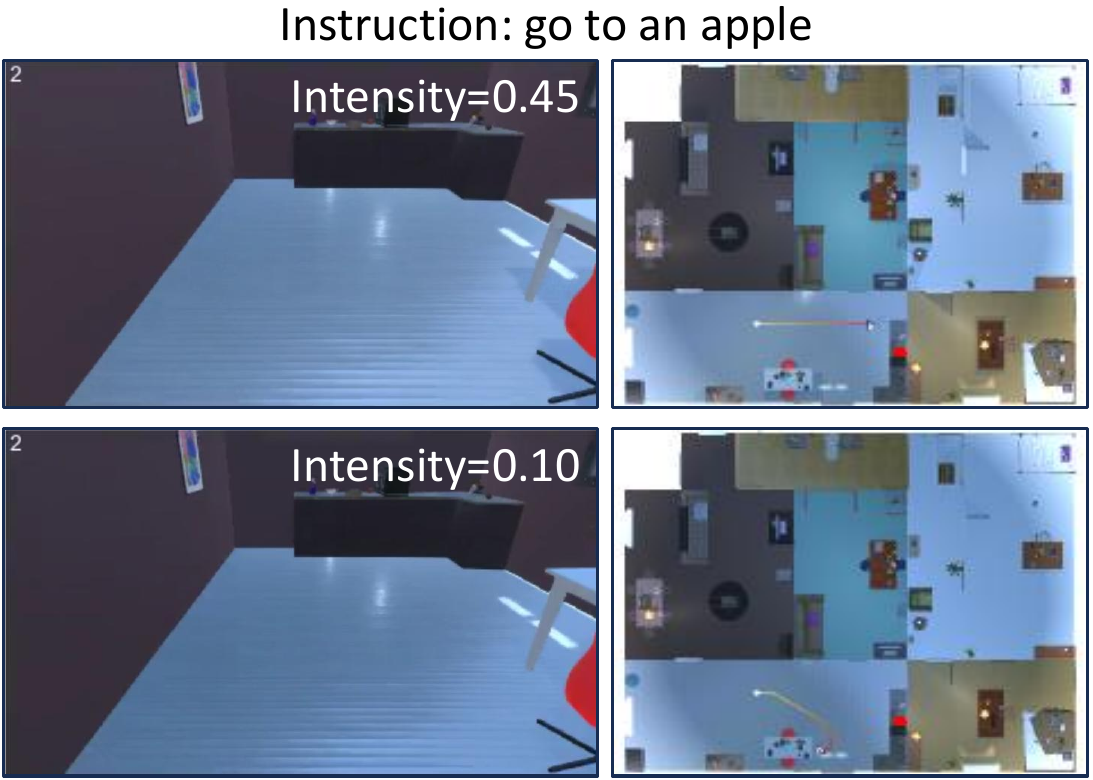}
        \caption{SPOC-ObjectNav}
        \label{fig:result_spoc_obj}
    \end{subfigure}
    \begin{subfigure}{0.311\linewidth}
        \centering
        \includegraphics[width=\linewidth]{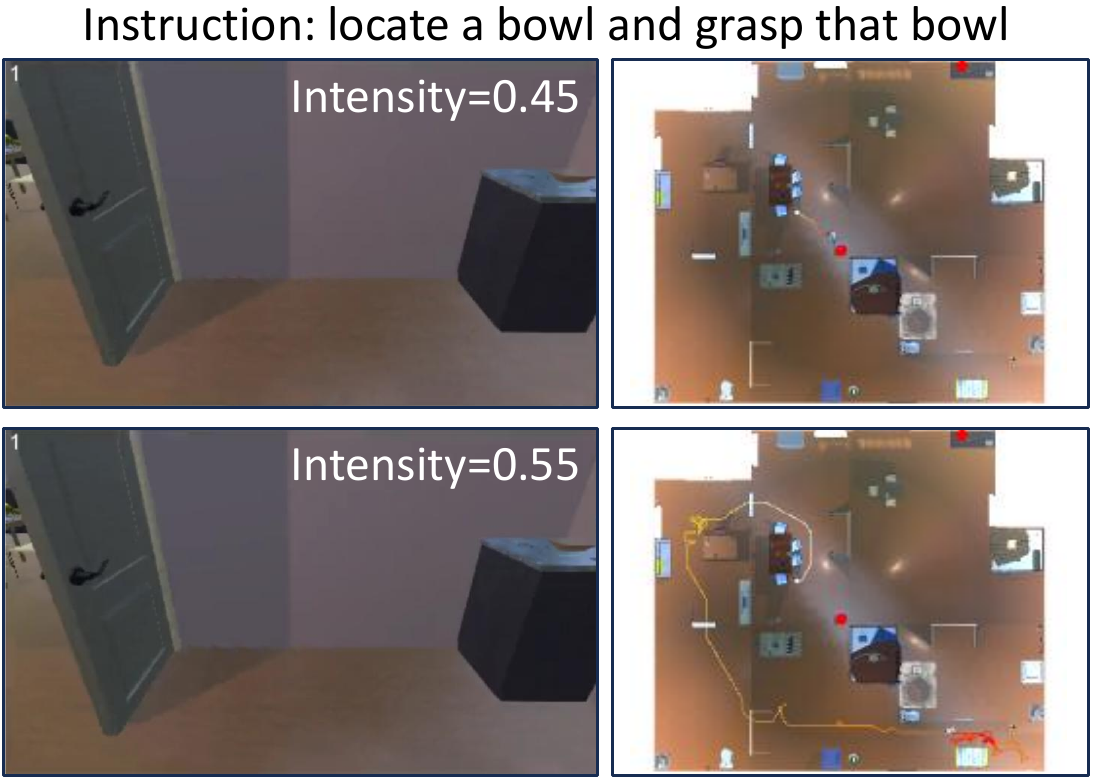}
        \caption{SPOC-Fetch}
        \label{fig:result_spoc_fetch}
    \end{subfigure}
    \begin{subfigure}{0.311\linewidth}
        \centering
        \includegraphics[width=\linewidth]{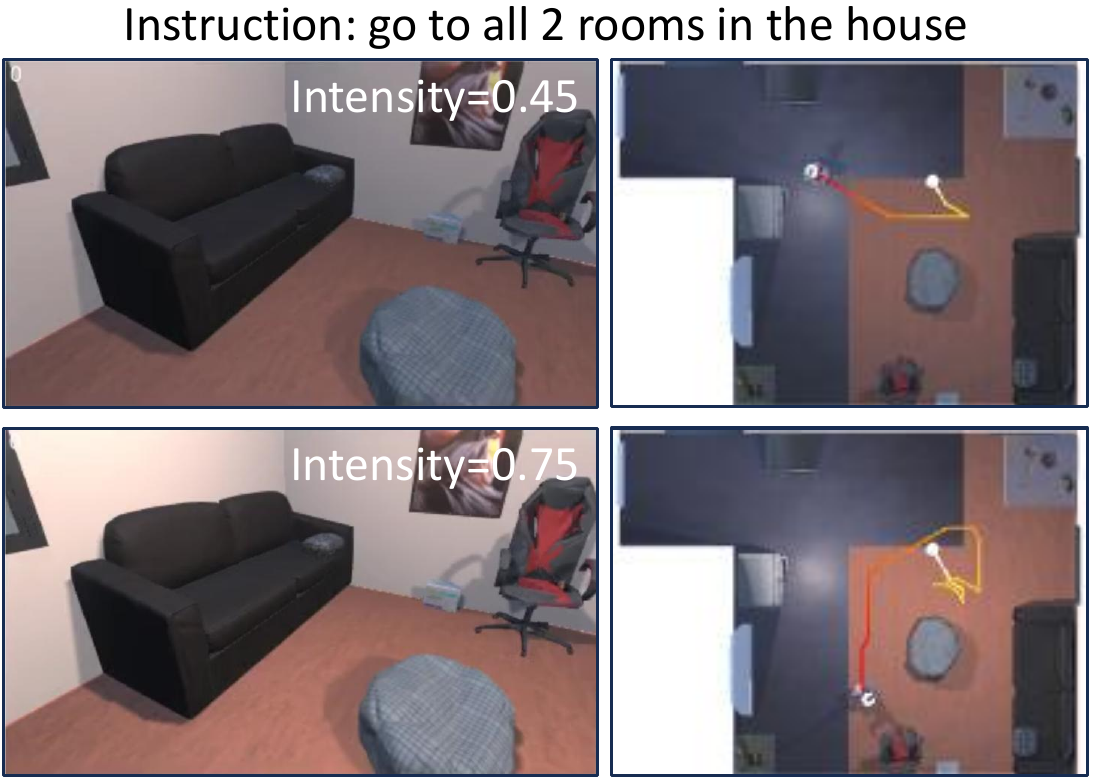}
        \caption{SPOC-RoomVisit}
        \label{fig:result_spoc_rv}
    \end{subfigure}
    
    \begin{subfigure}{0.05\linewidth}
        \centering
        \includegraphics[width=\linewidth]{figures/result-6-legend.pdf}
    \end{subfigure}
    \begin{subfigure}{0.311\linewidth}
        \centering
        \includegraphics[width=\linewidth]{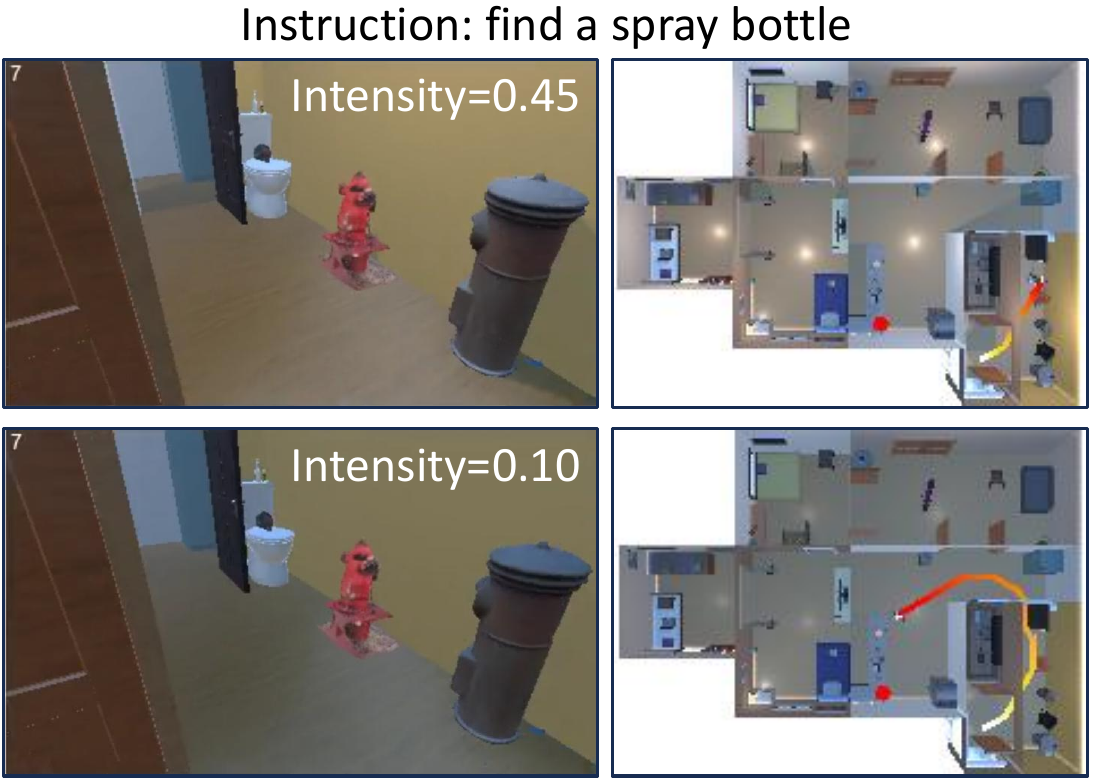}
        \caption{FLaRe-ObjectNav}
        \label{fig:result_flare_obj}
    \end{subfigure}
    \begin{subfigure}{0.311\linewidth}
        \centering
        \includegraphics[width=\linewidth]{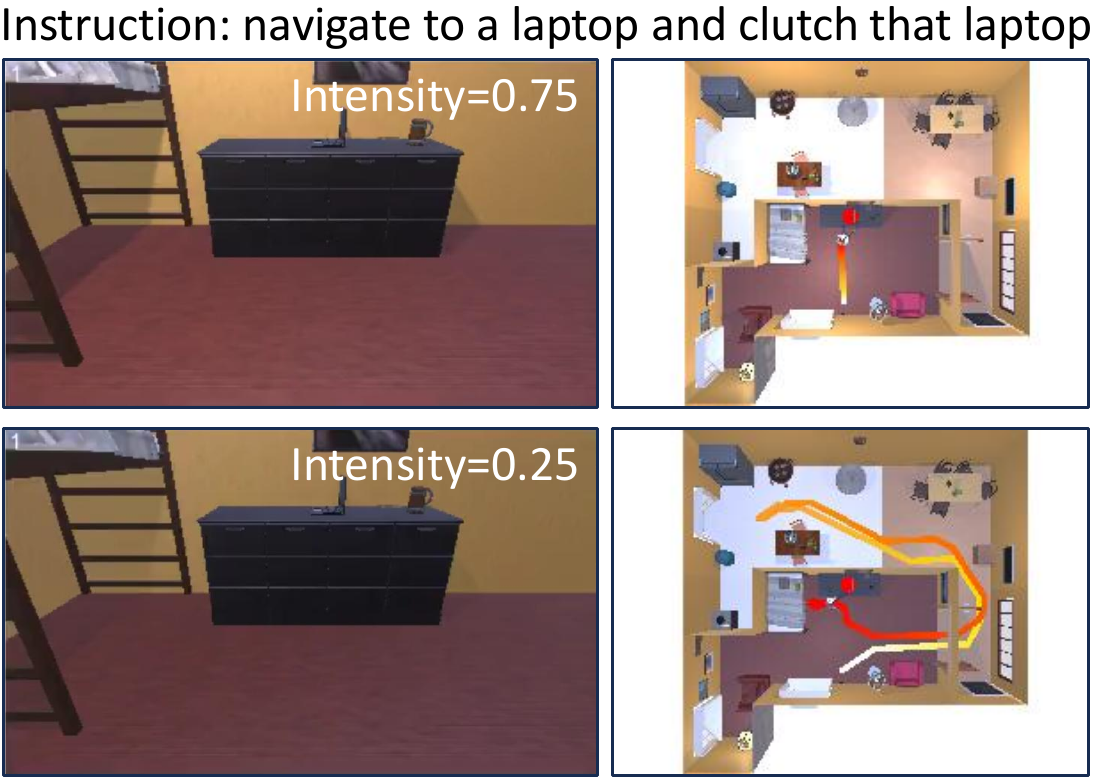}
        \caption{FLaRe-Fetch}
        \label{fig:result_flare_fetch}
    \end{subfigure}
    \begin{subfigure}{0.311\linewidth}
        \centering
        \includegraphics[width=\linewidth]{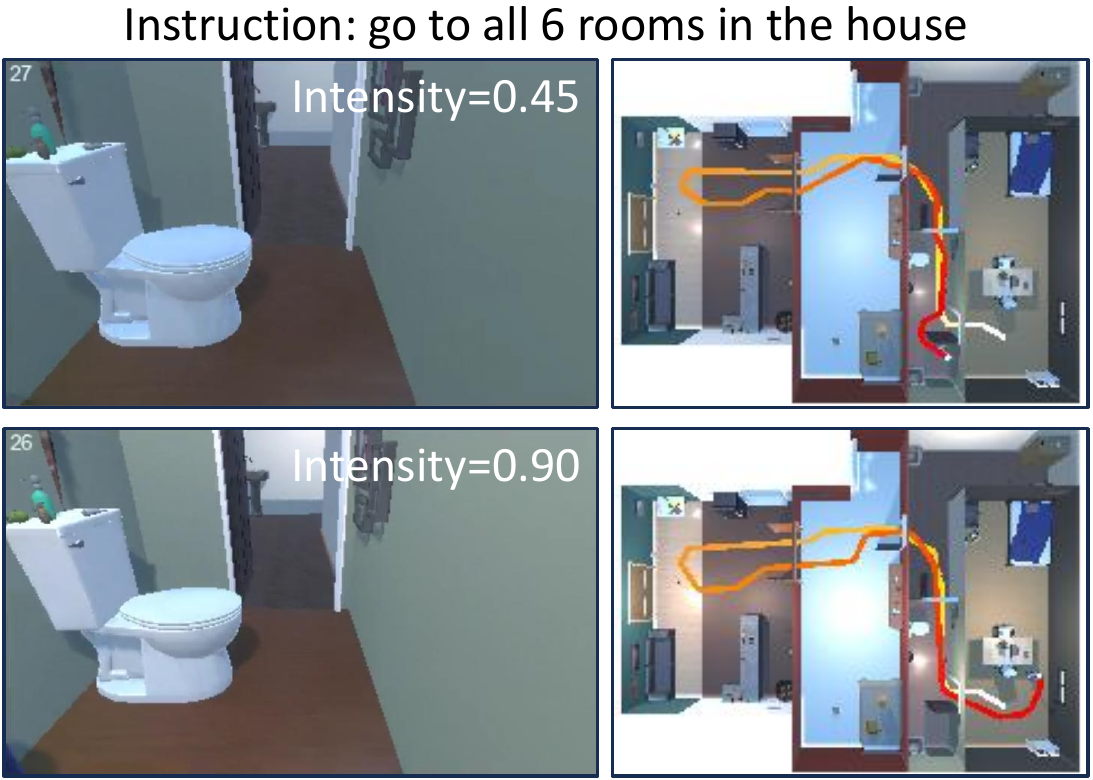}
        \caption{FLaRe-RoomVisit}
        \label{fig:result_flare_rv}
    \end{subfigure}
    \caption{
        Visualization of ego-centric visual observations and resulting trajectories under clean and attacked scenes by our proposed ILA. 
    }
    \label{fig:result}
\end{figure*}

%%%%%%%%%%%%%%%%%%%%%%%%%%%%%%%%%%%%%%%%%%%%%%%%%%%%%%%%%%%%%%%

\subsection{Attack Performance}
\label{ssec:main}

To evaluate the effectiveness of our method, we compare SILA and DILA against baselines on SPOC and FLaRe models across three navigation tasks. Table~\ref{tab:main_result} presents the attack performance of baseline methods and our proposed approaches.
Random Intensity achieves ASRs ranging from 12.20\% to 75.00\%, demonstrating that even unoptimized lighting variations can disrupt navigation. Texture-GA achieves substantially higher ASRs (37.57\% to 87.50\%), showing that adversarially optimized texture perturbations cause more significant disruption. Both baselines also increase EL, with Texture-GA showing more pronounced effects.
Our proposed SILA further enhances attack effectiveness, achieving ASRs from 47.27\% to 100\% and moderately increasing EL, indicating reduced navigation efficiency across episodes. This improvement stems from two key advantages: SILA leverages loss-guided optimization to identify more disruptive illumination configurations and employs global illumination changes that affect broader spatial regions, leading to more persistent disruption than localized texture perturbations.

Furthermore, incorporating DILA into SILA brings substantial improvements. The combined approach achieves ASRs ranging from 52.73\% to 100\%, with improvements of 5.46 to 35.85 percentage points across five task-model combinations (SPOC-Fetch already reaches 100\% with SILA alone). EL also increases substantially, often doubling (\eg, SPOC-ObjectNav: 131.07 to 234.27). By strategically triggering lighting switches at moments to increase heading deviation, DILA progressively steers agents off course, amplifying attack effectiveness even where static lighting perturbations show limited impact. 
Notably, our approach demonstrates strong generalization across diverse task types: from single-target localization (ObjectNav), to manipulation-requiring tasks (Fetch), to multi-room exploration (RoomVisit). This consistent effectiveness indicates that lighting-based attacks exploit fundamental visual perception vulnerabilities rather than task-specific weaknesses.

Figure~\ref{fig:result} presents representative examples comparing agent behavior in clean and attacked environments across different models and tasks. Each example shows ego-centric observations (left) and corresponding top-down trajectories (right). Our attack perturbs the agent's visual perception by adjusting lighting intensity. The examples illustrate how underlit scenes obscure spatial details while overlit scenes wash out critical features. The trajectory maps reveal that attacked agents deviate from optimal paths, failing to reach their goal locations.

%%%%%%%%%%%%%%%%%%%%%%%%%%%%%%%%%%%%%%%%%%%%%%%
\begin{table}[t]
\setlength{\tabcolsep}{1.5pt}
\centering
\caption{Ablation study for loss design in SILA and lighting switch triggering strategy in DILA. 
% We compare our timestep-weighted loss against two variants: one using final-step loss and another unweighted version.
}
\label{tab:ablation}
\resizebox{1.0\linewidth}{!}{
\begin{tabular}{|cc|ccc|cc|}
\hline
\multicolumn{2}{|c|}{\multirow{3}{*}{\textbf{ASR↑(\%)}}} & \multicolumn{3}{c|}{\textbf{Loss Design}} & \multicolumn{2}{c|}{\textbf{Switch Strategy}} \\
\cline{3-7}
\multicolumn{2}{|c|}{} & \textit{Final} & \textit{Unwei-} & \textit{Ours} & \textit{Random} & \textit{Ours} \\
\multicolumn{2}{|c|}{} & \textit{Step} & \textit{ghted} & \textit{(SILA)} & \textit{Trigger} & \textit{(DILA)}\\
\hline
\multirow{3}{*}{SPOC}
 & ObjectNav & 56.44 & 57.94 & \textbf{60.38} & 85.85 & \textbf{96.23} \\
 & Fetch & \textbf{100.00} & \textbf{100.00}  & \textbf{100.00} & \textbf{100.00} & \textbf{100.00}  \\
 & RoomVisit & 51.28 & 47.50 & \textbf{52.44} & 69.51 & \textbf{70.73}  \\
\hline
\multirow{3}{*}{FLaRe}
 & ObjectNav & 46.71 & 44.44 & \textbf{47.27} & 47.88 & \textbf{52.73} \\
 & Fetch & 66.99 & 64.42 & \textbf{68.81} & 74.31 & \textbf{93.58} \\
 & RoomVisit & 55.38 & 54.55 & \textbf{57.48} & 70.08 & \textbf{74.80} \\
\hline
\end{tabular}
}
\end{table}
%%%%%%%%%%%%%%%%%%%%%%%%%%%%%%%%%%%%%%

%%%%%%%%%%%%%%%%%%%%%%%%%%%%%%%%%%%%%%%%%%%%%%%%%%%%%%%%%%%%%%%%%
\subsection{Ablation Study}
\label{ssec:ablation}

To evaluate the effectiveness of our design choices, we conduct ablation studies on the timestep-weighted loss in SILA and the lighting switch triggering strategy in DILA.

\noindent\textbf{Effect of timestep-weighted loss in SILA.}
To evaluate the effectiveness of our loss design, we compare our timestep-weighted loss against two variants: a final-step loss that only considers the Euclidean distance between the agent's final position and the target, and an unweighted loss that treats all timesteps equally.
Table~\ref{tab:ablation} shows that our timestep-weighted loss consistently achieves the highest ASR across all six task-model combinations. Notably, on SPOC-Fetch, all three loss variants reach 100\% ASR, indicating this task's high vulnerability to lighting attacks. Our approach consistently outperforms both variants on the remaining five combinations, validating the effectiveness of trajectory-wide loss aggregation. Moreover, assigning higher weights to timesteps closer to the goal focuses optimization on critical navigation stages, leading to more effective attacks than considering only the distance of the final position and the destination or treating all timesteps uniformly.

\noindent\textbf{Effect of switch triggering strategy in DILA.}
To evaluate the effectiveness of our switch triggering strategy, we compare it against a random baseline that applies lighting switches at arbitrary timesteps while maintaining the same total number of switches per episode.
Table~\ref{tab:ablation} shows that our strategy consistently achieves higher ASR across all task-model combinations. These results demonstrate that our one-step lookahead surrogate effectively identifies critical moments for lighting switches. By estimating heading deviation at each timestep, our strategy selectively triggers switches when they more strongly deflect the agent from the goal direction, leading to more targeted disruptions than random switching. This validates that strategic timing of lighting perturbations is crucial for attack effectiveness.

%%%%%%%%%%%%%%%%%%%%%%%%%%%%%%%%%%%%%%%%%%%%%%%%%%%%%%%%%%%%%%%%%

\subsection{Parameter Discussion}
\label{ssec:parameters}

In this section, we investigate the impact of key parameters, including the intensity bounds and perturbation step size in SILA, and the number of lighting switches in DILA.

\noindent\textbf{SILA parameter analysis.} 
In our method, SILA searches over global lighting intensity to identify disruptive illumination configurations. This search is governed by two key parameters: intensity bounds that define the admissible range, and perturbation step size that controls the granularity of iterative updates. To analyze their impact, we evaluate nine configurations by pairing three intensity ranges ([0.0, 1.0], [0.0, 1.5], [0.0, 2.0]) with three step sizes (0.01, 0.05, 0.10) on SPOC-ObjectNav.
Table~\ref{tab:static_param_analysis} shows that ASR ranges from 50.94\% to 60.38\% across all settings.
Analyzing the results, we observe that moderate step sizes (0.05) outperform both small steps (0.01), which suffer from slow convergence, and large steps (0.10), which risk overshooting optimal values. Similarly, moderate lighting bounds ([0.0, 1.5]) achieve better performance than narrow ranges ([0.0, 1.0]), which limit search coverage, or excessively broad ranges ([0.0, 2.0]), which reduce optimization efficiency as effective configurations become sparser. These findings demonstrate that balancing search space coverage with optimization granularity is crucial for attack effectiveness. Therefore, we adopt lighting intensity bounds [0.0, 1.5] and perturbation step size 0.05 as the default SILA setting.

%%%%%%%%%%%%%%%%%%%%%%%%%%%%%%%%%%%%%%%%%%%%%%%%%%%%%%%%%%%%%%%%%
\begin{table}[t]
\centering
\caption{Attack performance of SILA under different combinations of intensity bounds and per-iteration perturbation step sizes. 
}
\label{tab:static_param_analysis}
\resizebox{\linewidth}{!}{
\begin{tabular}{|c|c|ccc|}
\hline
\multicolumn{2}{|c|}{\multirow{2}{*}{\textbf{ASR↑(\%)}}} & \multicolumn{3}{c|}{\textbf{Per-iteration Step Size}} \\
\cline{3-5}
\multicolumn{2}{|c|}{} & 0.01 & 0.05 & 0.10 \\
\hline
\multirow{3}{*}{Bounds} 
 & \text{[0.0, 1.0]} & 54.29 & 57.58 & 55.56 \\
 & \text{[0.0, 1.5]} & 50.94 & \textbf{60.38} & 58.25 \\
 & \text{[0.0, 2.0]} & 50.94 & 56.73 & 58.10 \\
\hline
\end{tabular}
}
\end{table}

%%%%%%%%%%%%%%%%%%%%%%%%%%%%%%%%%%%%%%%%%%%%%%%%%%%%%%%%%%%%%%%%%

%%%%%%%%%%%%%%%%%%%%%%%%%%%%%%%%%%%%%%%%%%%%%%%%%%%%%%%%%%%%%%%%%
\begin{table}[t]
\setlength{\tabcolsep}{4pt}
\centering
\caption{Attack performance of DILA under varying numbers of lighting switches. 
}
\label{tab:dynamic_param_analysis_switch}
\resizebox{\linewidth}{!}{
\begin{tabular}{|cc|cccc|}
\hline
\multicolumn{2}{|c|}{\multirow{2}{*}{\textbf{ASR↑(\%)}}} & \multicolumn{4}{c|}{\textbf{Number of Switches}} \\
\cline{3-6}
\multicolumn{2}{|c|}{} & 1 & 2 & 3 & Unlimited \\
\hline
\multirow{3}{*}{SPOC}
 & ObjectNav & 63.21 & 63.21 & 65.09 & \textbf{96.23} \\
 & Fetch & \textbf{100.00} & \textbf{100.00} & \textbf{100.00} & \textbf{100.00} \\
 & RoomVisit & 60.98  & 62.20 & 63.41 & \textbf{70.73} \\
\hline
\multirow{3}{*}{FLaRe}
 & ObjectNav & 48.48 & 47.27 & 47.88 & \textbf{52.73}\\
 & Fetch & 84.40 & 84.40 & 89.91 & \textbf{93.58} \\
 & RoomVisit & 68.50 & 70.08 & 70.08 & \textbf{74.80} \\
\hline
\end{tabular}
}
\end{table}

%%%%%%%%%%%%%%%%%%%%%%%%%%%%%%%%%%%%%%%%%%%%%%%%%%%%%%%%%%%%%%%%%

\noindent\textbf{DILA parameter analysis.}
In our method, DILA strategically triggers lighting switches at critical timesteps to amplify attack effectiveness. The key parameter is the number of switches per episode, which determines how frequently illumination changes disrupt the agent's visual perception. To analyze its impact, we evaluate four configurations: 1, 2, and 3 fixed switches per episode, as well as an unlimited mode that triggers switches at every eligible timestep identified by our surrogate function, on SPOC and FLaRe across three tasks.
Table~\ref{tab:dynamic_param_analysis_switch} shows a clear trend: ASR consistently increases with switching frequency, with the unlimited configuration achieving the highest performance across all task-model combinations. Increasing switching frequency improves attack effectiveness by disrupting the agent at multiple critical timesteps, leading to greater cumulative deviation from the goal. Although limited switches (1-3) achieve reasonable performance, the unlimited configuration maximizes disruption opportunities and consistently delivers optimal ASR. Therefore, we adopt the unlimited switch configuration as the default DILA setting.

%%%%%%%%%%%%%%%%%%%%%%%%%%%%%%%%%%%%%%%%%%%%%%%%%%%%%%%%%%%%%%%%%
\section{Conclusion}
\label{sec:conclusion}
In this work, we proposed a black-box adversarial paradigm that perturbs indoor lighting to evaluate the robustness of VLN agents under usual environmental attribute variations. The proposed method explores globally disruptive illumination configurations and strategically triggers lighting on/off to interfere with the agent's perception. Extensive experiments demonstrate that lighting-based attacks significantly degrade navigation performance across tasks and models. Future work will explore illumination-aware training and defense mechanisms for safer embodied AI systems.

\bibliographystyle{IEEEtran}
\bibliography{reference}

\end{document}